\documentclass[letterpaper, 10 pt, conference]{ieeeconf}  %

\IEEEoverridecommandlockouts                              %

\usepackage{amsmath} %
\usepackage{amssymb}  %
\usepackage{graphicx} %

\usepackage{booktabs}
\let\labelindent\relax
\usepackage{enumitem}

\usepackage[breaklinks,colorlinks]{hyperref}
\usepackage{siunitx}
\usepackage[ruled,vlined]{algorithm2e}
\usepackage{algorithmic}
\usepackage{microtype}
\usepackage{flushend}
\usepackage[hang,flushmargin]{footmisc}

\usepackage{url}
\usepackage[table, dvipsnames]{xcolor}
\makeatletter
\g@addto@macro{\endtabular}{\rowfont{}}%
\makeatother
\newcommand{\rowfonttype}{}%
\newcommand{\rowfont}[1]{%
\gdef\rowfonttype{#1}#1\ignorespaces%
}
\makeatother
\usepackage[export]{adjustbox}

\usepackage{pifont}%

\usepackage{siunitx}
\usepackage{cuted}
\usepackage{threeparttable}
\usepackage{multirow}
\usepackage{cite}
\usepackage{makecell}

\usepackage{tikz}
\usepackage{pgfplots}

\usepackage[resetlabels]{multibib}
\newcites{S}{References}

\usepackage{tabularx}
\newcolumntype{Y}{>{\centering\arraybackslash}X}
\newcolumntype{Z}{>{\raggedleft\arraybackslash}X}

\definecolor{dark-green}{RGB}{12,80,12}

\newcommand{\secref}[1]{\hyperref[#1]{Sec.~\ref*{#1}}}
\renewcommand{\eqref}[1]{\hyperref[#1]{Eq.~(\ref*{#1})}}
\newcommand{\figref}[1]{\hyperref[#1]{Fig.~\ref*{#1}}}
\newcommand{\tabref}[1]{\hyperref[#1]{Tab.~\ref*{#1}}}
\newcommand{\algref}[1]{\hyperref[#1]{Alg.~\ref*{#1}}}
\newcommand{\appref}[1]{Supplementary \secref{#1}}

\newcommand{\greyrule}{\arrayrulecolor{black!30}\midrule\arrayrulecolor{black}}

\newcommand{\para}[1]{\parskip=5pt\noindent\textit{#1}}
\newcommand{\ours}{MORE}

\newcommand{\website}{\url{https://more-model.cs.uni-freiburg.de}}

\newcommand{\ntwo}{N$^2$M$^2$}

\newcommand{\nrtasks}{81} %
\renewcommand{\baselinestretch}{0.985}

\title{\LARGE \bf
MORE: Mobile Manipulation Rearrangement Through \\ Grounded Language Reasoning
}

\author{Mohammad Mohammadi$^{1,2*}$, Daniel Honerkamp$^{1*}$, Martin Büchner$^{1*}$, Matteo Cassinelli$^{3*}$, Tim Welschehold$^{1}$, \\[0.4em] Fabien Despinoy$^{3}$, Igor Gilitschenski$^{2}$, and Abhinav Valada$^{1}$%
\thanks{$^{*}$ Equal contribution.}%
\thanks{$^{1}$ Department of Computer Science, University of Freiburg, Germany.}%
\thanks{$^{2}$ Department of Computer Science, University of Toronto, Canada.}%
\thanks{$^{3}$ Toyota Motor Europe.}%
\thanks{This work was funded by Toyota Motor Europe, an academic grant from NVIDIA, and the BrainLinks-BrainTools center of University of Freiburg. }%
}

\begin{document}

\maketitle
\thispagestyle{empty}
\pagestyle{empty}

\begin{abstract}
Autonomous long-horizon mobile manipulation encompasses a multitude of challenges, including scene dynamics, unexplored areas, and error recovery. Recent works have leveraged foundation models for scene-level robotic reasoning and planning. However, the performance of these methods degrades when dealing with a large number of objects and large-scale environments. To address these limitations, we propose \ours{}, a novel approach for enhancing the capabilities of language models to solve zero-shot mobile manipulation planning for rearrangement tasks. \ours{} leverages scene graphs to represent environments, incorporates instance differentiation, and introduces an active filtering scheme that extracts task-relevant subgraphs of object and region instances. These steps yield a bounded planning problem, effectively mitigating hallucinations and improving reliability. Additionally, we introduce several enhancements that enable planning across both indoor and outdoor environments. We evaluate \ours{} on \nrtasks{} diverse rearrangement tasks from the BEHAVIOR-1K benchmark, where it becomes the first approach to successfully solve a significant share of the benchmark, outperforming recent foundation model-based approaches. 
Furthermore, we demonstrate the capabilities of our approach in several complex real-world tasks, mimicking everyday activities. We make the code publicly available at \website{}.
\end{abstract}

\section{Introduction}
Recent studies have achieved notable advances in the completion of long-horizon robotic tasks in large-scale environments~\cite{rana2023sayplan, momallm24, shah2024bumble, schmalstieg2022learning}. This progress was largely fueled by recent breakthroughs in scene comprehension and the integration of foundation models. At the same time, evaluations are still limited to known environments~\cite{rana2023sayplan, shah2024bumble}, interactive search tasks~\cite{momallm24, schmalstieg2023learning}, or a series of hand-crafted tasks in specific real-world settings that lack reproducibility by the community~\cite{rana2023sayplan, shah2024bumble, zhi2024closed, nasiriany2024pivot, ichter2022do}.

\setlength{\tabcolsep}{1pt}

\begin{figure}[t]
	\centering
	\resizebox{1.0\linewidth}{!}{%
 \includegraphics[width=\linewidth,trim={0cm 6cm 0cm 0cm},clip,angle =0,valign=c]{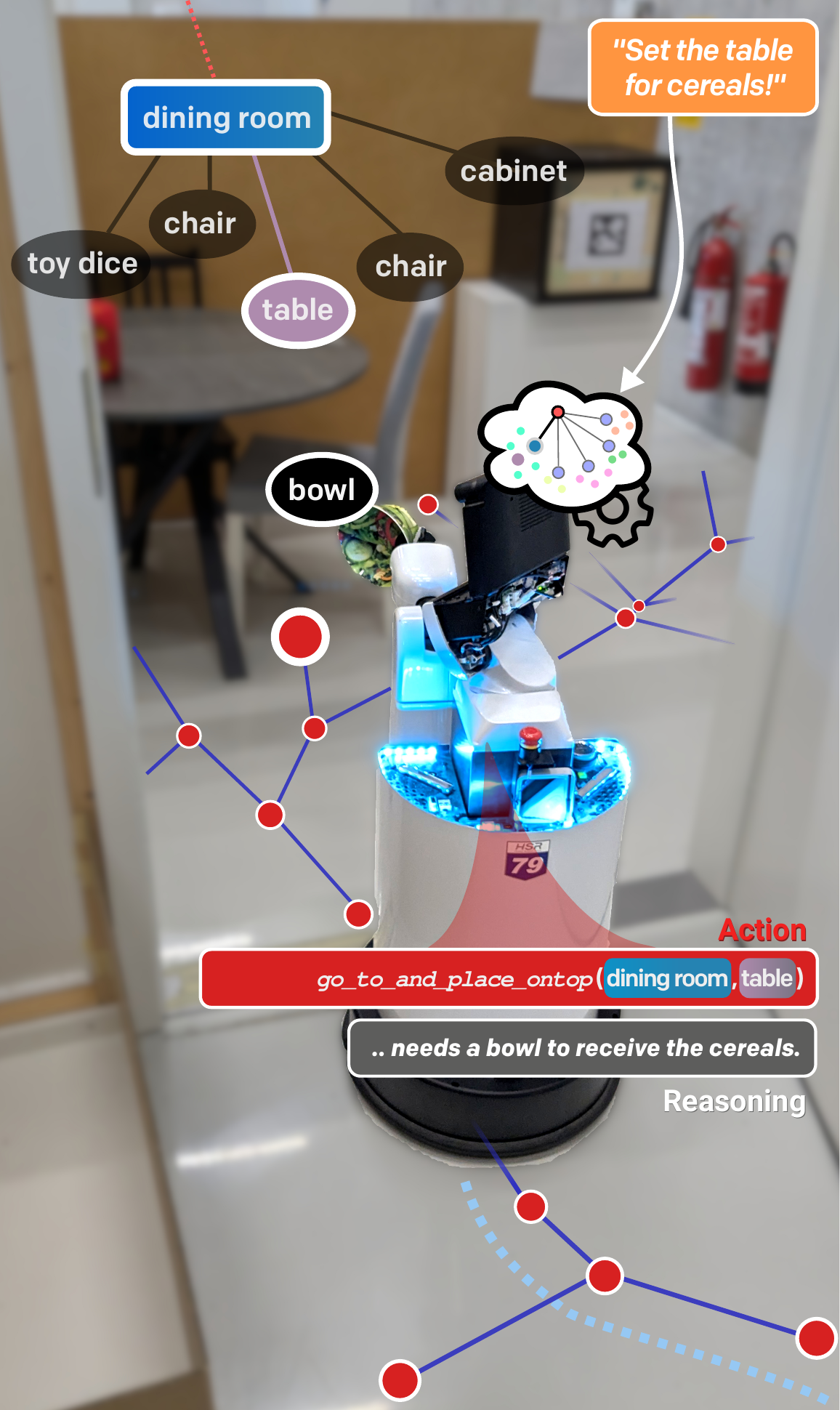}}
     \caption{We present \ours{}, an efficient model for the task of rearrangement through mobile manipulation.  We utilize 3D scene graphs as a logical scene representation manifold that is filtered to obtain task-relevant subgraphs.} 
  	\label{fig:teaser}
 \vspace{-0.4cm}
\end{figure}
\setlength{\tabcolsep}{6pt}

Previous research conducted within single-room environments has demonstrated the capability to accomplish a broad spectrum of tasks~\cite{shridhar2020alfred, nasiriany2024robocasa}. Nonetheless, in the context of mobile robotic manipulation, the scope of environments can expand to encompass entire apartments and outdoor areas, leading to an exponential increase in the number of objects and possible interactions, leading to exploding planning times~\cite{agia2022taskography} or hallucinations~\cite{rana2023sayplan}. Furthermore, assumptions of a priori known scene layouts and object locations do not account for environment changes or flexible deployment. These partial scene observability and unexplored areas complicate planning by requiring reasoning about the unknown.

In this work, we focus on generalizing rearrangement tasks with a mobile manipulator in a zero-shot manner, tackling both large-scale and unexplored environments spanning across indoor and outdoor spaces.
Building upon previous work in natural language-based interactive object search~\cite{momallm24}, we expand its application to encompass general everyday activities and introduce the first method capable of solving a substantial part of the BEHAVIOR-1K benchmark~\cite{li2023behavior}. We refine the benchmark by providing fully defined language task descriptions and implementing computationally efficient evaluations, thereby facilitating reproducible assessments across an extensive set of tasks.
Subsequently, we present a methodology that can operate effectively in unknown environments, both indoor and outdoor, and extends the reasoning to a large set of subpolicies. Our approach introduces novel solutions to integrate object and instance information and to identify the subscene pertinent to the current task, thereby maintaining the planning problem within manageable bounds and significantly mitigating hallucinations and distractions.

In our experiments, we conduct a comparative analysis of previous approaches based on Large Language Model (LLM) and Vision-Language Model (VLM) planners. Our experiments reveal substantial benefits from incorporating explicit memory, such as scene graphs, and employing our filtering approach to streamline the planning phase by minimizing the search space.

Our key contributions can be summarized as follows:
\begin{itemize}
    \item A novel scene representation that encompasses environments spanning across both indoor and outdoor areas.
    \item An efficient LLM-based planning framework that identifies the task-relevant subscenes and can execute instance-level actions in large-scale environments.
    \item The extension of the BEHAVIOR-1K benchmark with fully-specified language task descriptions, facilitating reproducible assessments of everyday activities.
    \item The first approach to complete a significant subset of BEHAVIOR-1K tasks, containing \nrtasks{} daily activities together with real-world evaluation of representative tasks in a multi-room apartment.
\end{itemize}

\section{Related Work}

\para{3D Scene Graphs and Spatial AI}: Robotic reasoning and planning struggle to scale effectively in large-scale environments when utilizing dense semantic representations~\cite{jatavallabhula2023conceptfusion,prasanna2024perception}. Instead, recent works break down large-scale indoor or outdoor scenes into hierarchical 3D scene graphs covering distinct concepts at various levels of granularity.~\cite{hughes2022hydra, werby23hovsg, greve2023curb, strader2024indoor}. While Hydra~\cite{hughes2022hydra} focuses on real-time execution, ConceptGraphs~\cite{gu2023conceptgraphs} and HOV-SG~\cite{werby23hovsg} investigate the incorporation of open-vocabulary vision-language features for language-grounded robot navigation. OrionNav~\cite{devarakonda2024orionnav} and MoMa-LLM~\cite{momallm24} further introduce frontier-based exploration stacks for dynamically populating the scene graph based on language-guided high-level policy calls. Orthogonally, KESU~\cite{gao2024enhancing} proposed constructing a knowledge base by employing vision-language foundation models to enhance spatial environment understanding. Clio~\cite{maggio2024clio} proposes to cluster task-agnostic object primitives based on task instructions to reduce scene complexity.
Related to the problem of scene granularity, Search3D~\cite{takmaz2024search3d} proposes an additional hierarchical level below the object primitives by breaking down class-agnostic object proposals into granular segments representing partial objects.
While still in its early stages, another branch of studies deals with dynamic updates of scene graphs through various input signals. Most works solely focus on the aspect of incorporating novel observations~\cite{momallm24,devarakonda2024orionnav}, whereas MM-3DSGU~\cite{olivastri2024multi} is the first general approach to also consider human interactions, robot actions, or language inputs ~\cite{olivastri2024multi}.
Compared to previous works, we propose a scene graph approach that spans both indoor and outdoor domains under a single combined representation and allows space partitioning through Voronoi decomposition.

\para{Vision-Language Models for Planning}: Foundation models have shown strong reasoning capabilities that provide accurate representations. Language models have been utilized to formulate a planning domain and establish a goal for a symbolic planner to execute planning within~\cite{birr2024autogpt, liu2024delta}. Coupled with scene graphs, LLMs have been able to solve long-horizon tasks~\cite{rana2023sayplan, rajvanshi2023saynav, momallm24}. 
Recently, VLMs have enabled finer-grained decision-making based on annotated images~\cite{wang2024grounding, nasiriany2024pivot}. Where COME relies on global semantically annotated object maps to enable longer-horizon reasoning~\cite{zhi2024closed}, OK-Robot~\cite{liu2024ok} only executes open-vocabulary pick-and-drop tasks. BUMBLE further extends to additional skills and includes visual memory~\cite{shah2024bumble}. Furthermore, fully end-to-end models have succeeded in performing tasks such as unloading and folding laundry. However, these models concentrate on confined environments~\cite{nasiriany2024pivot, wang2024grounding, zhi2024closed, liu2024ok, black2024pi_0} that do not require robust memory mechanisms,  rely on a priori known scenes~\cite{rana2023sayplan, shah2024bumble} or involve a restricted range of interactions related to search and exploration~\cite{momallm24, rajvanshi2023saynav}. %

In large environments, methods to reduce the size of the planning problem have been proposed by predicting the importance of objects through graph neural networks~\cite{silver2021planning}. Liu~\textit{et~al.}~\cite{liu2024delta} query an LLM to prune a fully known scene graph given a task description. Ray~\textit{et~al.}~\cite{ray2024task} identify rules for removing symbols from a task and motion planning (TAMP) navigation problem. In contrast to these methods, our work concentrates on formulating plans within unknown environments, thus necessitating an analysis of the significance of nodes in the context of exploration. EmbodiedRAG~\cite{booker2024embodiedrag} enables an LLM to envision task-relevant objects, subsequently formulating plans only with those objects, which correspond to the a priori knowledge.

\para{Embodied AI Benchmarks}: In the last years, numerous simulators and benchmarks have been introduced, including large and navigable scenes~\cite{habitat19iccv}. Recently, they have been extended to partially interactive scenes for rearrangement tasks, without doors or articulated objects~\cite{puig2024habitat, gan2021threedworld}, and often restricted to confined room-sized scenes~\cite{shridhar2020alfred, nasiriany2024robocasa}. Where iGibson~\cite{li2021igibson} provides 15 fully interactive apartment scenes, ProcThor procedurally generates large scenes, nonetheless with limited complexity~\cite{deitke2022procthor}. A number of recent works have shown capabilities on real-world hand-crafted tasks, complex to replicate and use for benchmarking~\cite{rana2023sayplan, shah2024bumble, zhi2024closed, nasiriany2024pivot, ichter2022do}. In an effort to enable evaluations that are both accessible and replicable, we have chosen BEHAVIOR-1K ~\cite{li2023behavior}, a dataset that augments the scenes from iGibson to encompass 50 varied daily task scenarios with near-photorealistic visualization. It offers an open and challenging testbed for long-horizon task planning in large environments that require strong capabilities of understanding and memory.

\section{Approach}
\setlength{\tabcolsep}{1pt}
\begin{figure*}[t]
	\centering
	\resizebox{\textwidth}{!}{%
 \includegraphics[width=\textwidth,trim={0cm 0cm 0cm 0cm},clip,angle =0,valign=c]{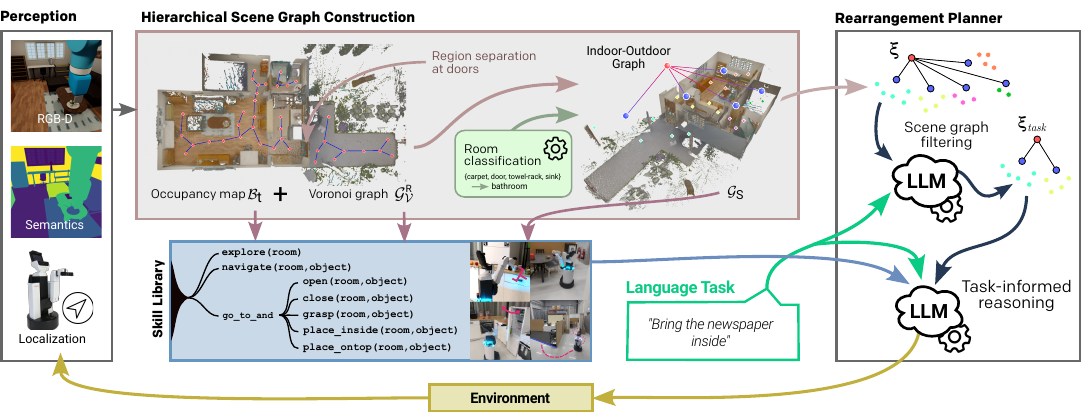}}

     \caption{Overview of \ours{}. Starting in an unexplored environment, we continuously construct a hierarchical scene graph of the environment based on RGB-D data and semantic segmentation. We first build a dense occupancy map and then extract a navigational Voronoi graph. This graph is separated at room borders based on door locations, then clustered and segmented into rooms and outdoor regions. The resulting scene graph is converted to natural language descriptions. To generate a bounded planning problem, we first filter the observed objects by task relevance and then use an LLM as a task planner in the resulting subgraph. The LLM orchestrates navigation, manipulation, and exploration subpolicies, which in turn result in an updated scene representation.}
  	\label{fig:overview}
   % \vspace{-0.3cm}
\end{figure*}
\setlength{\tabcolsep}{6pt}

\subsection{Problem Statement: Embodied Reasoning}\label{sec:task}

This work investigates embodied reasoning in large and unexplored environments. The agent is situated in an unknown environment and given a task description $g$ in natural language together with its proprioceptive sensor observations, consisting of a posed RGB-D frame $I_t$ and the current robot state $s_{robot}$. For computational simplicity, we assume accurate semantic perception by relying on ground truth semantics. To complete the task, the agent has to explore its surroundings, identify relevant objects given the task, and interact with the environment until all goal conditions are satisfied.

Although recent studies are centered on manually defined tasks executed under real-world conditions~\cite{rana2023sayplan, shah2024bumble, zhi2024closed, nasiriany2024pivot, ichter2022do}, these tasks frequently exhibit a high degree of specificity unique to each research endeavor and are often challenging to reproduce. This specificity has resulted in numerous approaches that are difficult to compare.
For this reason, we focus on one of the broadest benchmarks for everyday tasks, BEHAVIOR-1K~\cite{li2023behavior}. While this benchmark has been established for a considerable period, it remains unsolved. Its public availability and broad scope enable reproducible, large-scale evaluation of general task solving capabilities.

The BEHAVIOR-1K benchmark consists of 1,000 activities in over 50 scenes across both indoor and outdoor areas~\cite{li2023behavior}. Each task is defined by its associated objects as well as initial and final conditions in the predicate logic language BDDL~\cite{srivastava2022behavior}. The tasks are then procedurally initialized in the OmniGibson environment, built on top of the NVIDIA Isaac simulator. The environment samples the required task objects and places them in a randomly chosen scene such that they fulfill the initial task conditions. Following successful initialization, the agent, embodied as a Fetch robot, is then randomly placed in the scene.
We carefully analyzed both the task environments and definitions and provide the following benchmark modifications for practicality and reproducibility. We verified that all scenarios are computationally feasible in a reasonable amount of time.

We categorized the tasks into two groups: currently solvable and nearly unsolvable. Specifically, we isolate a subset of 163 tasks that necessitate interactions involving articulated objects and, in the broadest interpretation, entail rearrangements with \textit{inside} and \textit{onto} state transitions. We further exclude any task wherein the initialization process is unsuccessful. The final set includes \nrtasks{} tasks such as \texttt{packing picnic food into car} or \texttt{buy dog food} %
across 17 unique scenes.
Given that the original task descriptions frequently lack unique specificity regarding the desired task outcomes, we provide natural language descriptions derived from the BDDL goal conditions for the complete set of tasks. For instance, the natural language description for the first mentioned task is: \texttt{Put all the food in the bags and load the bags into the car.}
The enumeration of the tasks, along with their corresponding natural language descriptions, is accessible in \appref{sec:task_subset}.

Nevertheless, conducting evaluations across this extensive set of tasks is exceedingly time-consuming, owing to their prolonged duration and the constrained simulation speed within large environments. We address this problem by introducing significantly accelerated subpolicies. These policies follow the concept of ``magic actions'', removing the need for physical simulations and their associated steps. Although this methodology disables the precise execution of arm movements, it enables a reasonable simulation pace to evaluate high-level reasoning capabilities across a broader set of tasks compared to previous works. We make these subpolicies publicly available with our code. The implementation of these magic actions is detailed in \appref{sec:sim_environment}.

\subsection{\ours{} Model}\label{sec:ours}
We build on previous work that takes advantage of scene graphs and language-model based planners to reason and act in large environments~\cite{momallm24} and extend it to general tasks. Our model consists of three main components: a hierarchical scene graph representation, a high-level language-based planner, and subpolicies that control the robot. An overview is shown in \figref{fig:overview}.
\setlength{\tabcolsep}{1pt}
\begin{figure}[t]
	\centering
	\resizebox{\linewidth}{!}{%
 \includegraphics[width=\textwidth,trim={0cm 0cm 0cm 0cm},clip,angle =0,valign=c]{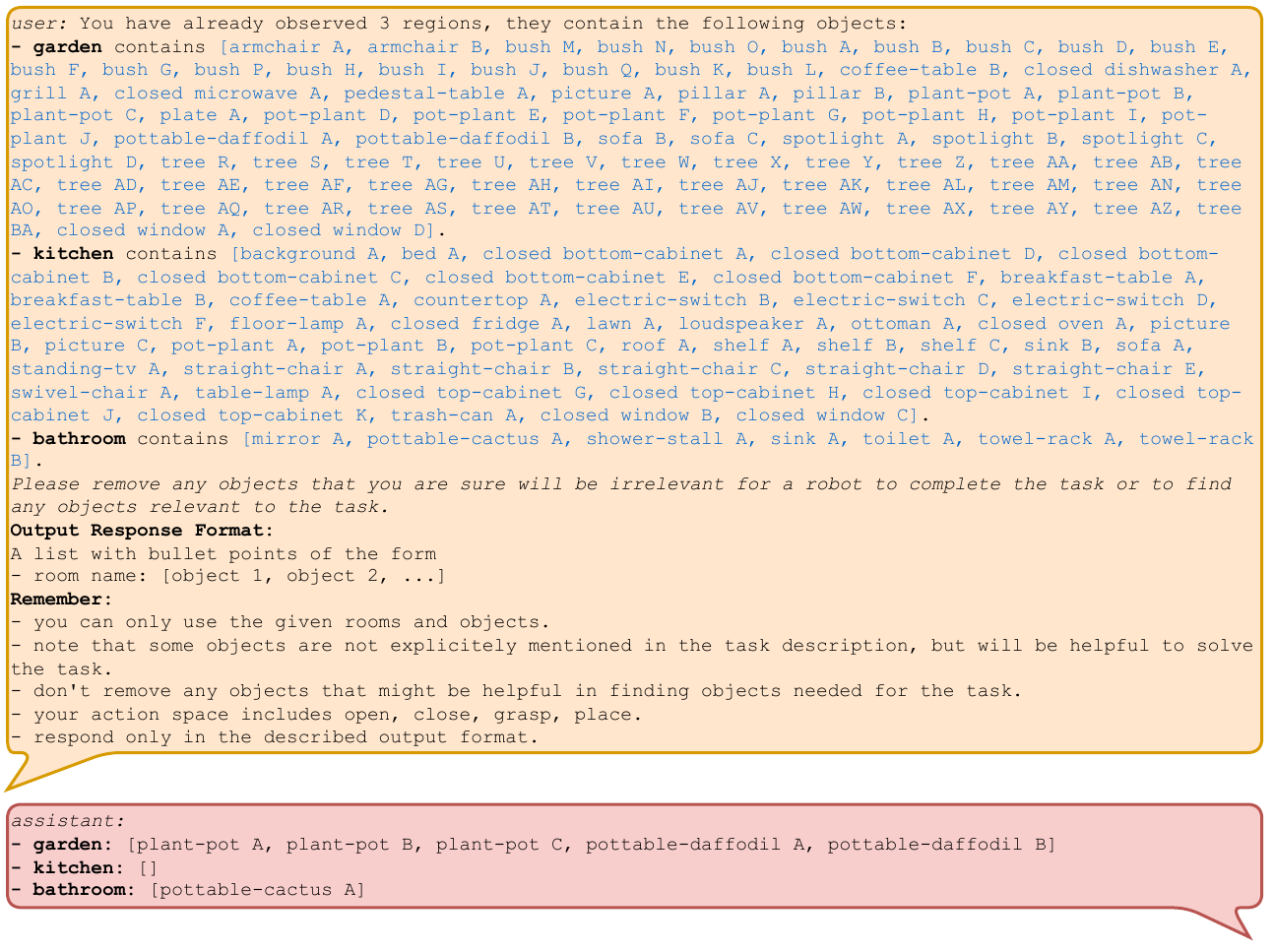}}
     \caption{Language-based scene graph filtering: We employ a filtering prompt that structurally represents all regions and respective objects of the unfiltered scene $\xi$. Next, we employ an LLM in order to identify which objects are task-relevant. Those are returned, yielding a task-informed sub-graph $\xi_{task}$.}
  	\label{fig:filter_prompt}
   % \vspace{-0.4cm}
\end{figure}
\setlength{\tabcolsep}{6pt}

\subsubsection{Scene Graph}
Given the requirement to act in combined indoor-outdoor scenes, we extend previous work to larger scenes. Similarly to MoMa-LLM~\cite{momallm24}, we employ Voronoi graphs to cover the navigational space. Although Voronoi graphs allow for efficient space partitioning, they quickly grow to an impractical number of nodes, which is exacerbated in the considered large-scale environments. In general, each morphological irregularity generates a substantial quantity of Voronoi nodes that do not contribute to functional utility. 
Therefore, we introduce a novel sparsification scheme operating on the navigational Voronoi graph, whereby we identify and extract nodes of degree two that fall within a specified distance of one another. We then replace the original edges with a single coarser edge. The resulting algorithm scales linearly in the number of vertices and is provided in \appref{app:voronoi_details}. 
While MoMa-LLM~\cite{momallm24} assigns all objects to the largest connected component of the Voronoi graph, this is no longer sufficient for areas that extend to the outdoors. A specific challenge is the observability of different regions of the scene through unnavigable windows. We instead assign objects directly to their closest component and only then reduce the observed scene graph to the component connected to the robot's current position. This eliminates unreachable regions until a path is discovered by opening a door, for instance.
Given these modifications, the proposed navigational graph can efficiently represent unified indoor-outdoor scenes that feature highly disparate node counts. In addition, we extend the room separation mechanism to an adaptive kernel width based on the observed door bounding boxes, thereby facilitating more robust segmentation of regions. Subsequently, we adhere to MoMa-LLM's strategy to assign objects to the separated Voronoi graph, based on both the viewpoint they are observed from and the language-model based room classification of each region.

\setlength{\tabcolsep}{1pt}
\begin{figure}[t]
	\centering
	\resizebox{\linewidth}{!}{%
 \includegraphics[width=\textwidth,trim={0cm 0cm 0cm 0cm},clip,angle =0,valign=c]{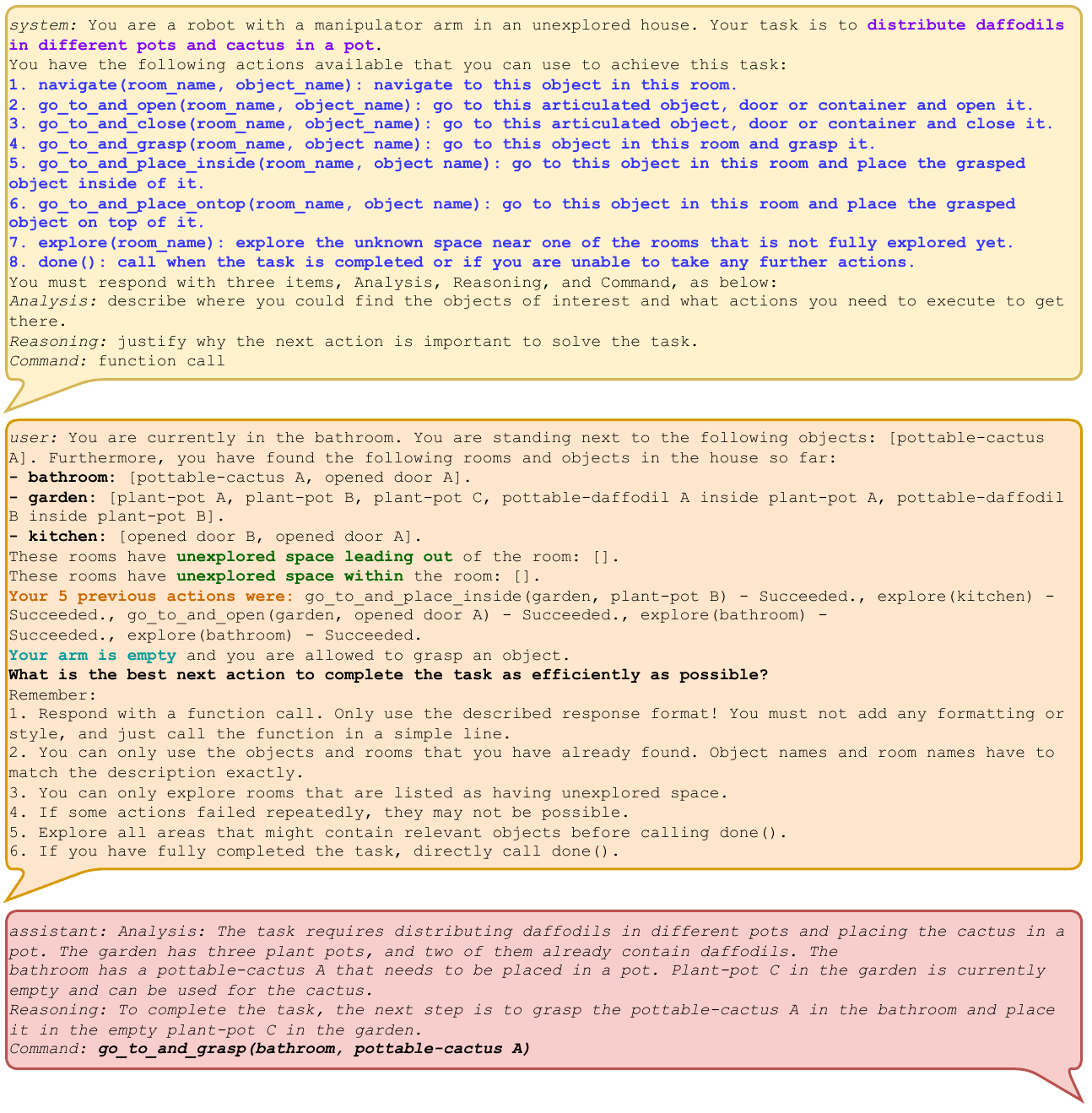}}
      % \vspace{-0.2cm}
     \caption{Language-based reasoning: Assuming a filtered sub graph $\xi_{task}$, we provide the LLM with the task description, a skill API, and the filtered set of regions and objects. Based on the high-level action chosen by the LLM, we orchestrate the next policy steps to be executed in order to fulfill the task at hand.}
  	\label{fig:reasoning_prompt}
    % \vspace{-0.3cm}
\end{figure}
\setlength{\tabcolsep}{6pt}

\subsubsection{High-level planning}
Given our perceived state of the scene, our aim is to take the action with the highest likelihood of progressing the task. We introduce a novel scene graph filtering mechanism and extend LLM-based reasoning to instances and attributes.

\para{Scene representation}: We extend prior research to encode the scene graph, including potential unobserved regions, into a structured language representation~\cite{momallm24}. Whereas previous research concentrated predominantly on class-specific object search, our objective is to generalize to a broader range of tasks. As a result, we represent the task relevance of object attributes and properties, even for multiple target object instances, in our scene graph representation.
Inferred object states or attributes, accessible through recent VLM methods~\cite{werby23hovsg}, are incorporated into their related description as supplementary attributes. However, given that objects can possess a large number of attributes, this can lead to a large increase of the scene description. We tackle this through the filtering described below.
As a task may require us to interact with a specific instance, we first uniquify each instance by assigning it an alphabetical ID and provide all instances to the LLM. Note that this principle is interchangeable with using specific properties as instance identifiers. However, alphabetical IDs enable fair comparisons across different methods.
The resulting scene description is depicted in \figref{fig:filter_prompt}. We specify each region and its corresponding objects including IDs and ask the LLM to remove object instances that are not relevant given the task as detailed in the following.

\para{Scene graph filtering}: Planning in the full environment is a highly challenging task, as considering all possible objects and interactions leads to either exploding planning times~\cite{agia2022taskography} or hallucinations~\cite{booker2024embodiedrag}.
For common everyday tasks, only a small subset of objects and areas in the scene are actually task relevant.
Therefore, we hypothesize that there exists a filtering function $f$ that maps from the full scene $\xi$ to the relevant subscene $\xi_{task}$, which is simpler to attain than the full task planning. Its aim is to ignore distracting scene entities.
In contrast to previous work~\cite{silver2021planning, ray2024task, liu2024delta}, we face two additional challenges: we are acting in unexplored environments, making not just task-specific objects relevant, but also objects that may inform us about the unexplored parts of the scene, and we are considering tasks that might require multiple instances of objects or instances with specific attributes.

We propose an LLM-based scene filtering for partially observed scenes. The LLM receives a structured natural language description of the scene together with the task instructions. It is then prompted to remove any objects that will be irrelevant for a robot to completing the task, or in finding any of the objects relevant for the task, and to respond with a formatted list of rooms and objects. An example is shown in \figref{fig:filter_prompt}.

\para{Task planning}: Given the identified subscene $\xi_{task}$, we conduct MPC-style task planning, outputting the next subpolicy to execute before replanning with the newest scene observation.  We instantiate this as another large language model call $a = l(\xi_{task}, s_{robot}, g, h)$, where $g$ is the task instruction, $h$ is an action history, including potential environment feedback from the subpolicy execution.
We extract two parts of information from the robot state $s_{robot}$: the room it is currently in,  and whether it is holding an object in its gripper. The action history $h$ consists of the previous five actions as well as all previously failed actions. The environment feedback $c$ consists of a binary indicator whether a subpolicy succeeded or failed, as well as easily obtainable feedback such as the gripper being full or requiring an articulated object to be opened before placing another object inside. While more information would be available in simulation, it is often not realistic to obtain such details in the real world.
An example prompt is shown in \figref{fig:reasoning_prompt}.

We parameterize object-centric subpolicies, consisting of \texttt{explore(room)}: explore the closest frontier within a specified room. \texttt{navigate(room, object)}: navigate to the vornoi node closest to this object via A$^*$-planner in the constructed occupancy map. \texttt{go\_to\_and\_\{open, close, grasp, place\_inside, place\_ontop\}(room, object)}: navigate to the closest Voronoi node, then execute a magic action to complete the interaction.
\texttt{done()}: complete the episode and evaluate success. These subpolicies on one hand incorporate often implied precursor actions such as moving to the object, and on the other hand are parameterized by object and room names. We found this important to match the reasoning level of the LLM. The LLM is provided with a short description of each subpolicy.

An episode terminates if the model calls \texttt{done()}, exceeds 50 high level steps or continues to act for more than five high-level steps after all goal conditions are completed.

\section{Experiments}\label{sec:experiments}

\subsection{Experimental Setup}
We evaluate our approach on the BEHAVIOR-1K benchmark~\cite{li2023behavior} with a Fetch robot and magic actions as detailed in \secref{sec:task}. For every model tested, we use GPT-4o as a language model. For MoMa-LLM and \ours, we use the simpler model GPT-3.5~Turbo for room classifications. 

\para{Metrics}: We compare the following statistics:\\
\textit{Success Rate (SR)}: share of tasks with all goal conditions completed and \texttt{done()} called correctly.. Note that this does not require the model call \texttt{done()} correctly after doing so.\\
\textit{Total Task Completion (TTC)}: share of tasks  with all goal conditions completed, irresp.\, of calling \texttt{done()} correctly.\\
\textit{Task Progression (TP)}: average share of completed goal conditions of all goal conditions.\\
\textit{Relative task progression (rTP)}: share of completed goal conditions that include observed task objects. Removing a component of ambiguity from tasks such as ``throw away \textit{all} boxes'' where it is unclear when to stop searching for further objects.

    \begin{table}
        \scriptsize
        \setlength{\tabcolsep}{5pt}
        \centering
        \caption{Results on \nrtasks{} tasks of the BEHAVIOR-1K Benchmark.}\label{tab:sim_experiments}
        \vspace{-.2cm}
        \begin{threeparttable}
        \begin{tabularx}{\linewidth}{l|cccc}
          \toprule
            Model & SR [\%] & TTC [\%] & TP [\%] & rTP [\%]\\%& \multirow{2}{*}{AUC-E}  & Object & Distance & Infeas.\\
             \midrule
            BUMBLE~\cite{shah2024bumble} & 11.3 &15.0 & 35.2  & 39.7\\
            BUMBLE~\cite{shah2024bumble} + filtering  & 15.7 & 20.5 & 43.7 & 48.8 \\
            MoMa-LLM~\cite{momallm24} & 19.7 & 32.1 & \underline{67.6}  & \underline{72.4}   \\
            MoMa-LLM~\cite{momallm24} + spatial relations & \underline{36.1} & \underline{41.2} & 64.6  & 67.4   \\
            \ours{} & \textbf{48.1} & \textbf{50.6} & \textbf{70.1} & \textbf{80.1} \\
          \bottomrule
        \end{tabularx}
        \begin{tablenotes}[para,flushleft]
           \footnotesize      
           The reported numbers are averaged across all evaluated tasks. We report the success rate (SR), the average task progression (TP), the relative task progression (rTP), and the TTC denoting the total task completion rate.
         \end{tablenotes}
       \end{threeparttable}
    \end{table}

\setlength{\tabcolsep}{3pt}
\renewcommand{\arraystretch}{1}
\begin{figure}
	\centering
    \footnotesize
    \setlength{\tabcolsep}{0.05cm}%
    {\renewcommand{\arraystretch}{1}%
    \resizebox{\linewidth}{!}{%
    \begin{tabular}{ccc}
  		\includegraphics[width=0.5\textwidth,clip,angle=0]{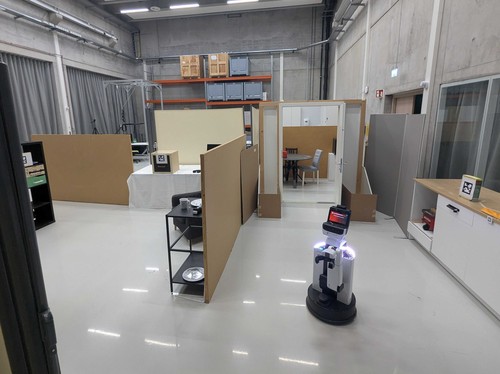} & 
        \includegraphics[width=0.5\textwidth,clip,angle=0]{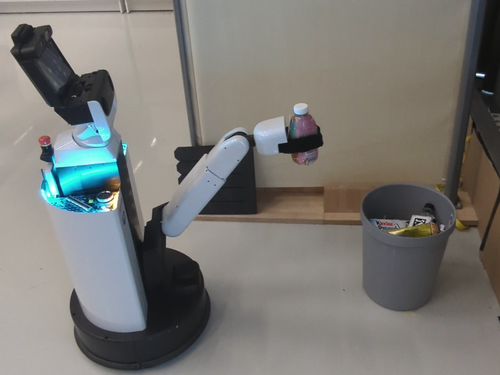} & \includegraphics[width=0.5\textwidth,clip,angle=0]{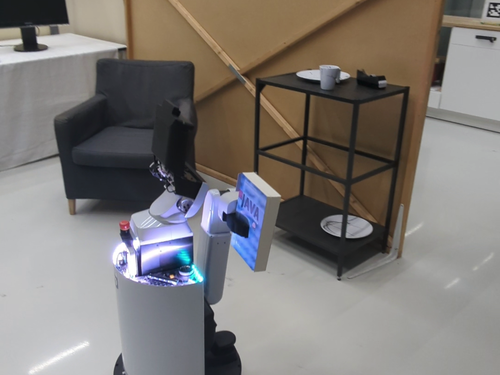}\\\includegraphics[width=0.5\textwidth,clip,angle=0]{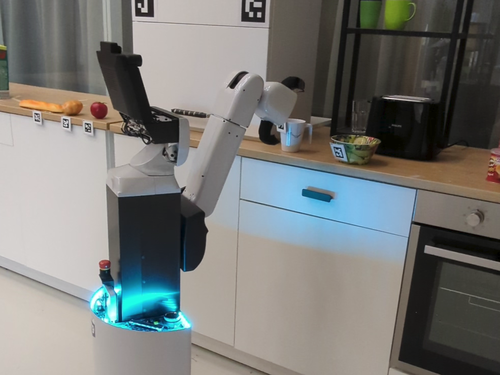} & \includegraphics[width=0.5\textwidth,clip,angle=0]{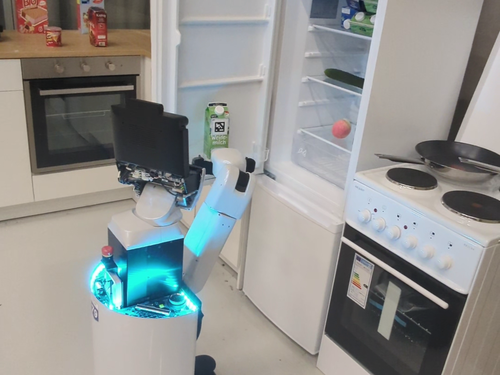} &
        \includegraphics[width=0.5\textwidth,clip,angle=0]{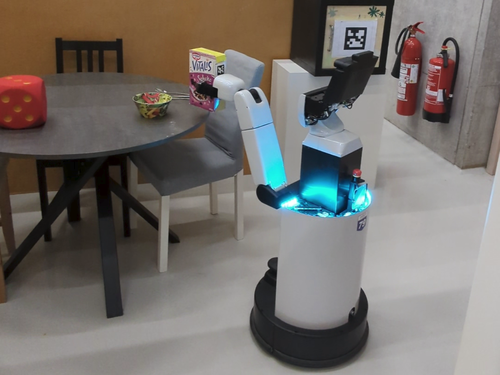} \\
    \end{tabular}}
    }
	\caption{Real-world experiments. From top left to bottom right: overview of the environment, throwing away a plastic bottle, fetching a book, grasping a bowl, taking milk out of the fridge, placing cereals on the dinner table.}
  	\label{fig:real_world}
    % \vspace{-0.3cm}    
\end{figure}

\setlength{\tabcolsep}{6pt}
\renewcommand{\arraystretch}{1}

    \begin{table*}
        \scriptsize
        \centering
        \caption{Real-world Evaluation}
        \vspace{-.2cm}
        \label{tab:real_world}%
        \begin{threeparttable}
        \begin{tabularx}{\linewidth}{p{2.5cm}p{3.6cm}p{0.8cm}Y}
          \toprule
            \centering Task & \centering Goal Conditions & \centering Completed & Reasoning \\
          \midrule
            \centering Set the table for cereals.       & \centering Place bowl, cereals and milk from fridge on dinner table & \centering 3/3 & Placed bowl and cereals on table; opened fridge to search for milk, found it, and placed it on table, then called done.\\
            \greyrule
            \centering Clean up the empty bottles and cans.                &  \centering Find two bottles and one can and put in the bin & \centering 3/3 & Found and threw away all objects. Continued to search kitchen and living room. Then failed to call done, stopped manually when base controller disconnected.\\
            \greyrule
            \centering Fetch a book and store it in the dinning room. & \centering Find book and place in storage box or cabinet in dining room & \centering 1/1 & Explored all rooms, found and grasped book, drove back to dining room and placed in storage box. Due to gripper opening too late book fell outside box. Then called done.\\
          \bottomrule
        \end{tabularx}
       \end{threeparttable}
       \vspace{-0.5cm}
    \end{table*}

\para{Baselines}: We compare \ours{} against a range of different methods detailed in the following.\\
\textit{MoMa-LLM}~\cite{momallm24}: conceptual predecessor of this work, encoding a scene graph into structure language form for an LLM planner, which we extend to our set of subpolicies.\\
\textit{MoMa-LLM}~\cite{momallm24}\textit{ + spatial relations}: the above model, equipped with an improved scene representation including spatial relations among observed objects, same as \ours{}.\\
\textit{BUMBLE}~\cite{shah2024bumble}: a current VLM-based reasoning method that provides a visual short- and long-term memory together with the current, annotated view to a VLM model to select subpolicies. We adapt the model to have access to the same ground-truth semantic perception as \ours, replace its original skill set with our subpolicies and  adapt the prompt descriptions accordingly. Note that BUBMLE does not employ a structured scene representation. %
Additional details are provided in \appref{sec:bumble_details}.\\ 
\textit{BUMBLE~\cite{shah2024bumble} + filtering}: the above model, extended with our scene filtering function, applied to the objects that get annotated in the current view.

\subsection{Results}

The results across the identified set of \nrtasks{} tasks are reported in \tabref{tab:sim_experiments}. We find that BUMBLE is able to complete a significant share of tasks. However, the landmark-based memory does not scale efficiently to very large environments, as this requires a very large number of frames. Furthermore, it does not allow reacting to changes in the scenes easily (which would require updated images). While the VLM reasoning can take into account fine-grained spatial relationships, it struggles with cluttered scenes that result in heavily annotated images, which in turn leads to incorrect selections or hallucinations. Accordingly, we observe an increase in performance when incorporating our filtering in this model. Nevertheless, the other issues persist. One advantage of MoMa-LLM and \ours{} over BUMBLE is that they base action selection on a comprehensive representation of the entire scene, rather than relying solely on the current view and the histories. This approach allows MoMa-LLM and \ours{} to achieve superior results compared to BUMBLE.
MoMa-LLM achieves a success rate of 19.7\%, though completing all conditions in 32.1\% of cases.
We found that it struggles with the large number of objects and fails to differentiate instances.
MoMa-LLM + spatial relations improves the differentiation of objects, enhancing performance and achieving a 36.1\% success rate underlining the effectiveness of our scene representation.

\ours{} achieves a success rate of 48.1\% and in over 50\% of all tasks completes all goal conditions. The average share of overall completed task conditions is even higher at over 70\%. Nonetheless, the model may face challenges in the following cases: (i) numerous constraints - although the model demonstrates high progress (rTP), it might overlook a few constraints and find a partial solution, (ii) task ambiguity - the task description might not provide the number of relevant objects, and the model struggles to determine when to stop searching, causing a gap between TP and rTP, (iii) simulator failures - spawning objects in unreachable areas, and getting stuck in dead spots due to opening articulated objects.\\
We conducted our experiments on a cluster consisting of 4 nodes, each equipped with 8 NVIDIA GeForce RTX 2080 Ti GPUs. We achieved an average completion time of 1 hour and 27 minutes per task.

% \vspace{-0.3cm}
\subsection{Real-World Experiments}\label{sec:real_world}

We construct an apartment consisting of three rooms, kitchen, living room, dining room, and deploy our approach on the Toyota HSR robot, with an omnidirectional base and 5-DoF arm. We follow previous work in assuming accurate semantic perception by revealing a pre-annotated semantic map based on the areas observed by the depth camera of the robot. We use the ROS Navstack for navigation and manipulate objects with the \ntwo{} mobile manipulation policy~\cite{honerkamp2021learning, honerkamp2023learning} based on single end-effector demonstrations per object, relative to an AR marker pose. For place poses, we infer the closest point to the robot on the selected object to place down objects, then generate linear end-effector motions to that pose.\looseness=-1

We evaluate our approach qualitatively on three challenging tasks. The results are reported in \tabref{tab:real_world} and shown in \figref{fig:real_world} as well as the accompanying video.
We find that our approach transfers well to the real world, completing all required goal conditions for all tasks, including complex interactions such as fetching the milk from the closed fridge. A small failure in opening the gripper led a book to fall outside the container. The largest issue was task ambiguity, leading to indeterminism whether the cleaning task is completed or if the agent should continue to search for further bottles. This reflects a general issue of abstract task definitions.
Additionally, we find long LLM inference times to cause a strain on evaluations. While these are restricted to high-level reasoning at a very low control frequency, they still accumulate to a significant fraction of runtime.

\section{Conclusion}
We introduced \ours{}, a model capable of diverse rearrangement tasks across very large environments.
We generate fully specified task descriptions for a large subset of the BEHAVIOR-1K tasks as well as a set of policies that allow computationally feasible evaluation. 
We construct scalable scene graphs across indoor/outdoor environments to subsequently identify the subscene and instance information directly related to the task. This enabled the efficient orchestration of object-centric subpolicies through a large language model.
In extensive experiments, we find our approach to outperform both previous LLM-based models and VLM-based approaches and, for the first time, complete a significant subset of the BEHAVIOR-1K benchmark. Finally, we demonstrated that our approach successfully transfers to real-world scenarios.
In future work, we aim to investigate the integration of VLM models for local fine-grained decision-making as well as further approaches for ambiguity resolution, either through human interaction, if available, or explicit reasoning about the desired task context.

\footnotesize
\vspace{-0.3em}
\bibliographystyle{IEEEtran}
\bibliography{bibliography.bib}

\clearpage
\renewcommand{\baselinestretch}{1}
\setlength{\belowcaptionskip}{0pt}

\begin{strip}
\begin{center}
\vspace{-5ex}
\textbf{\LARGE \bf
MORE: Mobile Manipulation Rearrangement Through \\ Grounded Language Reasoning
} \\
\vspace{3ex}

\Large{\bf- Supplementary Material -}\\
\vspace{0.4cm}
\normalsize{Mohammad Mohammadi$^{1,2*}$, Daniel Honerkamp$^{1*}$, Martin Büchner$^{1*}$, Matteo Cassinelli$^{3*}$, Tim Welschehold$^{1}$, \\[0.4em] Fabien Despinoy$^{3}$, Igor Gilitschenski$^{2}$, Abhinav Valada$^{1}$}\\
\end{center}
\end{strip}
\setcounter{section}{0}
\setcounter{equation}{0}
\setcounter{figure}{0}
\setcounter{table}{0}
\setcounter{page}{1}
\makeatletter

\renewcommand{\thesection}{S.\arabic{section}}
\renewcommand{\thesubsection}{S.\arabic{subsection}}
\renewcommand{\thetable}{S.\arabic{table}}
\renewcommand{\thefigure}{S.\arabic{figure}}

\let\thefootnote\relax\footnote{$^*$These authors contributed equally.\\$^1$Department of Computer Science, University of Freiburg, Germany.\\$^{2}$ Department of Computer Science, University of Toronto, Canada.\\$^3$Toyota Motor Europe (TME)\\
Project page: \website{}
}%

\normalsize 
In this supplementary material, we provide additional details on the generated task descriptions and simulation setup. We furthermore provide details on the scene graph construction and the implementation details of the BUMBLE baseline. We also provide additional details on the real world subpolicies.

\section{Task Subset}\label{sec:task_subset}
\tabref{tab:task_descriptions} presents the tasks utilized to evaluate our method. The first column lists the activity names, each corresponding to a task from the BEHAVIOR-1K dataset. The second column provides our new task descriptions given to the agent. These descriptions ensure that all task relevant details are included for the goal conditions provided by the BEHAVIOR-1K benchmark. For example, if goal conditions are specific to a certain object instance in the scene, the task description entails this detail. For tasks requiring interaction with specific object instances in the scene, the variable is preceded by a dollar sign \$.
During task initialization, variables preceded by a dollar sign \$ will be substituted with the corresponding object names.

\begin{table*}[ht]
        \scriptsize
        \setlength{\tabcolsep}{10pt}
        \centering
        \label{tab:task_descriptions}
        \caption{Task names \& task description for \nrtasks{} BEHAVIOR-1K tasks}
        \vspace{-0.2cm}
        \renewcommand{\arraystretch}{0.8}
        \begin{threeparttable}
        \begin{tabularx}{\linewidth}{l|l}
          \toprule
            Task name & Task description \\
             \midrule
putting\_away\_tools & put tools in toolbox. Wrenches must be in the same toolbox. Also screwdrivers must be in the same toolbox \\
\greyrule
attach\_a\_camera\_to\_a\_tripod & put the digital camera on top the tripod \\
\greyrule
sorting\_vegetables & distribute chard and bok choy in one bowl, leek and artichoke in another bowl, and sweet corns in another one \\
\greyrule
setting\_up\_room\_for\_games & put games and dices on \$table.n.02\_1 \\
\greyrule
recycling\_office\_papers & throw all legal documents in \$recycling\_bin.n.01\_1 \\
\greyrule
shopping\_at\_warehouse\_stores & put the cream cheese, watermelon, pomegranate, baguette and corn flakes on checkout \\
\greyrule
put\_together\_a\_scrapping\_tool\_kit & put the tools in the toolbox \\
\greyrule
putting\_dishes\_away\_after\_cleaning & put all the plates in \$cabinet.n.01\_1. In total there are 8 plates \\
\greyrule
fold\_a\_tortilla & put \$tortilla.n.01\_1 on \$plate.n.04\_1 and \$tortilla.n.01\_2 on \$tortilla.n.01\_1 \\
\greyrule
putting\_shopping\_away & store dairy and meat in \$electric\_refrigerator.n.01\_1, and other food and oil in \$cabinet.n.01\_1 \\
\greyrule
remove\_sod & put the sods on top of the wheelbarrow \\
\greyrule
distributing\_groceries\_at\_food\_bank & put milk, pasta, apple juice and canned for in each box \\
\greyrule
buy\_eggs & put \$egg.n.02\_1 and money on the checkout \\
\greyrule
packing\_fishing\_gear\_into\_car & load all fishing stuff into the car \\
\greyrule
hoe\_weeds & put weeds in \$ashcan.n.01\_1 \\
\greyrule
loading\_the\_car & put \$bag.n.06\_1, \$sack.n.01\_1 and \$laptop.n.01\_1 in the car \\
\greyrule
buy\_a\_good\_avocado & put the avocado and money on the checkout \\
\greyrule
packing\_documents\_into\_car & load the documents and the book into the car \\
\greyrule
pack\_a\_pencil\_case & put the pencil, pen, eraser and shear in the pencil box \\
\greyrule
packing\_car\_for\_trip & put sunglasses, money and laptop in suitcases, and load all suitcases into the car \\
\greyrule
packing\_art\_supplies\_into\_car & put the pencil and markers in the suitcase, and put the suitcase in the car \\
\greyrule
store\_beer & store all beer bottles in \$electric\_refrigerator.n.01\_1 \\
\greyrule
moving\_stuff\_to\_storage & put ice skates and bowling balls in one carton while the painting and textbooks in the other one \\
\greyrule
outfit\_a\_basic\_toolbox & put all the tools in the tool box \\
\greyrule
organizing\_boxes\_in\_garage & put balls, plates and saucepan in boxes \\
\greyrule
organizing\_office\_documents & put legal docs in a folder. Then, put post-it on the folder, and put the folder in \$cabinet.n.01\_1 \\
\greyrule
packing\_books\_into\_car & load all the books into the car and make sure suitcase is not inside the car \\
\greyrule
buy\_boxes\_for\_packing & put the cartons and money on the checkout \\
\greyrule
cleaning\_parks & put bottles in \$recycling\_bin.n.01\_1 \\
\greyrule
buy\_dog\_food & put the dog food and money on the checkout \\
\greyrule
buying\_groceries & \makecell{put milk, apple juice and prawn in one bag, and peanut butter, brown rice and banana in another bag. \\ put all the money in \$cash\_register.n.01\_1} \\
\greyrule
buy\_a\_air\_conditioner & put the air conditioner and money on the checkout \\
\greyrule
buy\_a\_keg & put the beer barrel and credit card on the checkout \\
\greyrule
buy\_pet\_food\_for\_less & put the canned foods and dog foods on checkout and all money in \$cash\_register.n.01\_1 \\
\greyrule
clearing\_the\_table\_after\_dinner & put bowls, cups and bottles in \$bucket.n.01\_1 \\
\greyrule
buy\_alcohol & put the wine, vodka, beer and money on the checkout \\
\greyrule
buying\_groceries\_for\_a\_feast & put all the groceries in a shopping card, and money in \$cash\_register.n.01\_1 \\
\greyrule
stacking\_wood & stack all logs on top of each other and put them on \$table.n.02\_1 \\
\greyrule
cleaning\_up\_branches\_and\_twigs & put branches in \$ashcan.n.01\_1 \\
\greyrule
buy\_food\_for\_a\_party & put all pasta, milk, apple juice, and cake on the checkout. Put all the money in \$cash\_register.n.01\_1 \\
\greyrule
buy\_candle\_making\_supplies & put candle sticks, paraffin and money on the checkout \\
\greyrule
putting\_out\_condiments & put the knifes, pickles, spaghetti sauce and all the bottles on \$breakfast\_table.n.01\_1 \\
\greyrule
returning\_videotapes\_to\_store & put DVDs in \$shelf.n.01\_1 \\
\greyrule
putting\_leftovers\_away & store all lasagna and spaghetti sauces in \$electric\_refrigerator.n.01\_1 \\
\greyrule
boxing\_books\_up\_for\_storage & put all the books in a carton \\
\greyrule
assembling\_gift\_baskets & put a cheese, candle, cookie and bow in each basket \\
\greyrule
re\_shelving\_library\_books & \makecell{put \$book.n.02\_1, \$book.n.02\_2, \$book.n.02\_3, and \$book.n.02\_4, \$book.n.02\_5, \$book.n.02\_6, \$book.n.02\_7 \\ and \$book.n.02\_8 on \$shelf.n.01\_1}
\\
\greyrule
sorting\_bottles\_cans\_and\_paper & distribute water bottles in one bucket, cans in another bucket, and magazines and newspapers in another one \\
\greyrule
preparing\_lunch\_box & store all the food and drink in \$packing\_box.n.02\_1 \\
\greyrule
bringing\_newspaper\_in & put the newspaper on \$coffee\_table.n.01\_1 \\
\greyrule
put\_togethera\_basic\_pruning\_kit & put the pruner and the shears in the toolbox \\
\greyrule
bag\_groceries & put egg, apple, orange juice and canned food in a bag \\
\greyrule
bringing\_water & put bottles in \$cabinet.n.01\_1 \\
\greyrule
preparing\_food\_for\_a\_fundraiser & distribute the bottles in cartons. Put the food on tupperware and then put tupperware in cartons \\
\bottomrule

\end{tabularx}

       \end{threeparttable}
    \end{table*}

\section{Simulation Environment} \label{sec:sim_environment}
\subsection{Modifications}
\begin{itemize}
    \item We close exterior doors and filter them out of the scenes as they lead to empty space.
    \item We take non-physical steps as the robot size cannot be adjusted, and many collisions between the robot and objects in narrow areas are unavoidable and would otherwise not allow the robot to complete certain tasks.
    \item We skip scene \texttt{restaurant diner} as its area is divided into 2 disconnected regions, and the scene \texttt{school gym} because of floors with different heights.
\end{itemize}
\subsection{Perception}
The robot in simulation is equipped with an RGB-D camera with a resolution of $256 \times 256$ pixels. As the focus of this work
is on decision making, we abstract from imperfect perception
and assume access to ground truth instances and semantic
segmentation from the simulator. For a realistic detection range,
we restrict all sensors (depth, semantics) to a maximum range
of $5$~m. We furthermore assume accurate detection of
whether an articulated object is open or closed. Moreover, all inner objects within an articulated object become visible to the robot once the outer object is opened. We construct all maps at a resolution of $0.075$ m and
detect the floor, carpet, lawn and driveway categories as free space.

\subsection{Execution} 
All trajectories start by fully turning around in-place to initialize the scene graph.

Low-level action space: The subpolicies act in a low-level action space consisting of the following actions:
\begin{itemize}
    \item move forward by \SI{7.5}{\cm}
    \item turn-left by up to \SI{35}{\radian}
    \item turn-right by up to \SI{35}{\radian}
    \item explore a region: the agent navigates to the closest frontier point of that region.
    \item open articulated object: if the arm is empty and the object is closed, the agent opens the object.
    \item close articulated object: if the arm is empty and the object is opened, the agent closes the object.
    \item grasp an object: if the arm is empty, and the agent grasps the object
    \item place an object relative to another object: the agent puts the object \{inside, on top\} of another object. If placing an object inside, the target object must be open.
    \item done: end the episode and evaluate the success
\end{itemize}

To execute any action in \{open, close, grasp, place\}, the robot must first navigate to a position near the target object. Specifically, it samples points along a fixed-radius circle around the object and moves to the closest sampled position. 

\begin{algorithm}[ht]
\caption{Voronoi Graph Sparsification} \label{alg:sparsification}
\begin{algorithmic}[1]
\REQUIRE Voronoi Graph $V$ with nodes $N$ and edges $E$, threshold value $c$.
\ENSURE Simplified Voronoi Graph

\STATE Let $E_2$ be the set of nodes in $N$ with a degree of 2
\FOR{each node $x \in E_2$}
    \STATE Let $v$ and $u$ be the neighbors of $x$
    \STATE Let $w_1$ and $w_2$ be the weights of edges $(v, x)$ and $(u, x)$ respectively
    \IF{$w_1 + w_2 < c$}
        \STATE Remove node $x$ from $V$
        \STATE Remove edges $(v, x)$ and $(u, x)$ from $V$
        \STATE Add edge $(u, v)$ with weight $w_1 + w_2$ to $V$
    \ENDIF
\ENDFOR

\end{algorithmic}
\end{algorithm}

\section{Voronoi Graph}\label{app:voronoi_details}
\subsection{Creation}
The agent perceives a sequence of posed RGB-D frames, $\{I_0, \dots, I_t\}$, including semantic maps of the environment. Using the camera's intrinsic parameters, the depth maps are converted into point clouds. These points are then transformed into a global coordinate frame and arranged on a 3D voxel grid, $M_t$. 
Since we address an interactive problem, our map is dynamically updated to incorporate newly explored areas and changes due to manipulated objects. 

To determine obstacle positions, walls, and free space, we extract the highest occupied entry per stixel in $M_t$. This value may be negative in cases where objects like swimming pools are present. These entries are then used to generate a two-dimensional bird’s-eye-view (BEV) occupancy map, $B_t$, by marking all occupied positions except those classified as free space, $F_t$. The latter represents the navigable area that is used in the exploration.

Similar to MoMa-LLM~\citeS{momallm24}, we construct a navigational graph, $G_V$ from the dense occupancy map. First, we inflate $B_t$ using a Euclidean Signed Distance Field (ESDF) formulation for robustness, while setting the free space coordinates defined in $F_t$ to zero. Based on this, we compute a Generalized Voronoi Diagram (GVD), which consists of a set of points, $V$, which maintain equal clearance to the nearest obstacles derived from $B_t$. We exclude all nodes that are on obstacles or do not reside within $B_t$. Given the boundaries of the GVD, we construct edges $E$ among $V$ and obtain our navigational Voronoi graph, $G_V = (V, E)$.

\subsection{Sparsification}
We employ the Voronoi graph as a proxy to extract the closest area to an object or detect frontier points in unexplored areas. However, the graph grows rapidly in large-scale environments. Therefore, any large process on the graph becomes computationally infeasible. In this case, each morphological irregularity consists of a vast number of Voronoi nodes that lie close to each other and do not add any functionality. We propose an iterative sparsification algorithm, described in \algref{alg:sparsification}, which reduces the complexity of the graph while preserving its structure.

Our algorithm removes redundant nodes with a degree of two if their neighbors are sufficiently close. The node is eliminated, and a direct edge is introduced between its neighbors. Note that the removal of one node does not change the degree of any other node, and after each removal step, exactly one degree-two node is eliminated. The algorithm processes each degree-two node only once. The final graph is minimal—i.e., no further degree-two nodes can be removed. This approach runs in a computational complexity of $O(|V|)$.

\subsection{Region Separation at Doors}
A scene is divided into regions when no navigable path exists between navigable regions, most commonly due to closed doors or gates. To extract the Voronoi graph for each region separately, we first remove all edges whose nodes belong to different regions. To achieve this, we treat all observed doors as separation gates and remove all edges that cross doors. Mathematically, we construct a 2D plane and fit a Gaussian kernel to each door, where the center of the kernel aligns with the center of the door. The Gaussian kernel diameters are adjusted according to the door's bounding box dimensions. Each door and its associated Gaussian kernel also share the same orientation.

Next, we evaluate each edge by integrating the values of its points in the 2D plane. If this integral exceeds a threshold, it indicates that the edge passes through a Gaussian kernel, that is, a door, implying that its nodes belong to different regions. As a result, when these edges are removed, it ensures a clear segmentation of the scene.

\section{Scene Graph}
Similar to MoMa-LLM~\citeS{momallm24}, we employ a hierarchical 3D scene graph with three levels of abstraction: root, regions, and objects. Consequently, our approach requires (i) distinguishing different regions within the environment and (ii) assigning each object to its corresponding region.

To detect different regions, we use the region separation method described in \appref{app:voronoi_details}. Due to partial observability, a single region (e.g., a room or garden) in the environment may be divided into multiple connected components within the Voronoi graph, each covering a sub-region. These sub-regions are merged once the exploration of the region is complete.

For each observed object $O$, we define $v_p(O)$ as the robot's viewpoint that is currently the closest to the object. We identify the nodes in the Voronoi graph that are in close proximity to the object and its viewpoint, denoted as $S_O$ and $S_{v_p(O)}$, respectively. 

For every pair $(v, u)$, where $v \in S_O$ and $u \in S_{v_p(O)}$, we determine the optimal solution to the following minimization problem:

\begin{equation}
\min_{v, u} \quad d_V(v, u) + d_E(O, v)^{\lambda} + d_E(v_p, u),
\end{equation}

where $d_V$ represents the shortest path length in the Voronoi graph, and $d_E$ denotes the Euclidean distance in 2D space. By weighting the distance to the object with an exponent of $\lambda = 1.3$, we prioritize selecting nodes that are closer to the object.  If $v$ and $u$ belong to different connected components in the Voronoi graph, the shortest path length $d_V(v, u)$ is considered infinite. Once the optimal pair $(v, u)$ is found, the object $O$ is assigned to the connectivity component that contains both $v$ and $u$.

After constructing the scene graph, the objects within each region are provided to an LLM to predict the region's label (e.g., kitchen, bathroom, etc.). Additionally, we determine the robot's current region and remove all unreachable regions and their associated objects from the scene graph.

\section{BUMBLE Implementation}\label{sec:bumble_details}
In this section, we will introduce the modifications we have made to BUMBLE~\citeS{shah2024bumble} to adapt it to the BEHAVIOR-1K tasks.
All implementation details provided below refer to the latest version of the author's repository at the time of writing\footnote{\url{https://github.com/UT-Austin-RobIn/BUMBLE}, commit \textit{4ff5ac0}.}.
BUMBLE uses Grounded SAM (GSAM)~\citeS{ren2024grounded} to obtain a segmentation mask of non-background objects. This is achieved by prompting GSAM with the query '\textit{all objects}'. 
For a fair comparison, we replaced GSAM with the simulator's ground truth information. Specifically, we obtained the segmentation mask and excluded background objects such as the floor and walls.
As previously described in section~\ref{sec:experiments}, we integrated the same set of subpolicies are used for all models for BUMBLE.

In the original BUMBLE pipeline, the agent could directly interact only with objects visible in the current frame, eliminating the need to store information about previously seen objects.
However, for the tasks tackled in this work, the agent may need to interact with previously discovered objects.
For this reason, we integrated a new buffer containing previously acquired frames. These frames are continuously sampled during action execution.

Whenever the agent intends to interact with an object, i.e., when performing \texttt{navigate} or \texttt{go\_to\_and\_$^*$} subpolicies, it must first select an image from the buffer containing the desired target object.
The frame selection process is implemented based on BUMBLE's \texttt{goto\_landmark} skill.
The agent is provided with the current subtask to achieve, a list of frames, each associated with a landmark. As we found BUMBLE prone to hallucinations, we additionally provide it with the list of objects within each image, which we found to reduce the problem to some degree. The agent then selects the landmark corresponding to the most suitable frame. 

Subsequently, the agent must choose one object from the selected frame.
This follows BUMBLE's \texttt{navigate\_to\_point\_on\_ground} skill. 
The agent is provided with the subtask to achieve, a description of all the objects in the selected frame, and the annotated frame. For annotating the image, we followed the public BUMBLE code.

The original work does not integrate a subpolicy for exploring the environment, relying instead on hardcoded positions within the tested buildings, which the robot could navigate between using the \texttt{goto\_landmark} skill.
We incorporate an exploration subpolicy following the same approach of the \texttt{goto\_landmark} skill, but, instead of using hardcoded positions, we provide the VLM with frames from the buffer where the unexplored areas are visible and a description of all the objects in each frame. The unexplored areas are extracted from the scene graph.
The agent subsequently chooses the image that depicts the areas to be explored. If multiple areas to explore are present in an image, the closest one is selected.

In BUMBLE, the feedback received after executing an action must be related to the subtask generated by the agent.
For the subpolicies \texttt{navigate} and \texttt{go\_to\_and\_$^*$} we use the same feedback as for our method.
For the more abstract exploration subpolicy, feedback is generated by the Visual Language Model (VLM). The VLM is tasked with determining whether the subtask was accomplished during the exploration. It is provided with the frames acquired during the exploration. Again, beyond BUMBLE's original implementation, we additionally include the list of objects in the frame to reduce halucinations.

\section{Real World Implementation}\label{sec:real_world_details}

\setlength{\tabcolsep}{1pt}
\begin{figure}[t]
	\centering
	\resizebox{\linewidth}{!}{%
 \includegraphics[width=\textwidth,trim={0cm 0cm 0cm 0cm},clip,angle =0,valign=c]{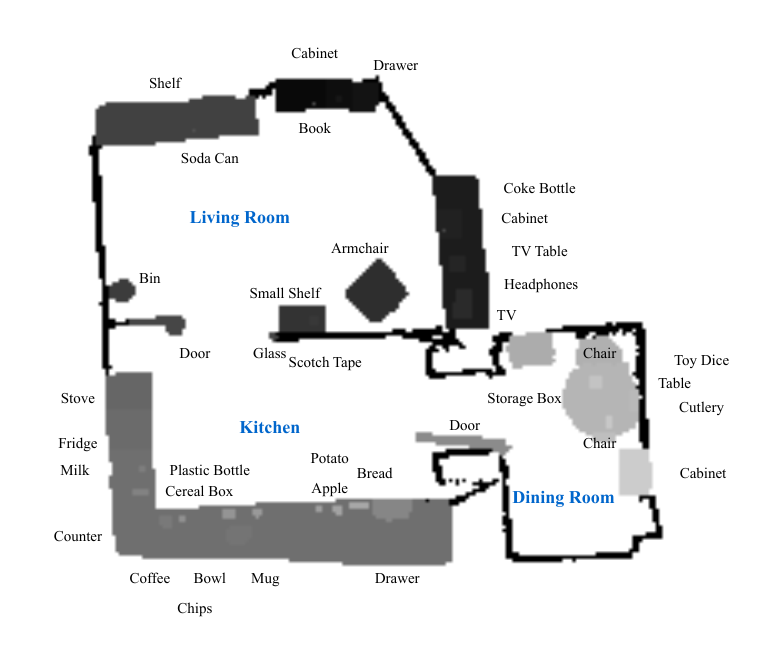}}
     \caption{Map of the real world environment. Object positions are shown in different intensities and annotated with their category name.}
  	\label{fig:map}
   % \vspace{-0.4cm}
\end{figure}
\setlength{\tabcolsep}{6pt}

\subsection{Map}
The real world environment spans three rooms, furnished as kitchen, living room and dining room with six articulated, openable objects and 38 different objects overall. A map of the environment is shown in \figref{fig:map}.

\subsection{Subpolicies}
As discussed in \secref{sec:real_world}, we simulate accurate semantic perception by annotating a recorded 2D map of the environment with semantic labels. We then use the depth camera of the robot to identify the explored area, and reveal the according parts of the map to robot. The recorded map is also used for LiDAR based localization. This allows us to isolate the factors relevant to the decision-making part of problem.

\texttt{Navigate}: the agent uses the default omnidirectional ROS Navstack provided by the robot's manufacturer. Obstacles are avoided based on the static map, inflated by \SI{0.25}{\meter} as well as dynamic obstacles perceived with the robot's LiDAR and depth camera mounted on the head.

\textit{Manipulation:} All manipulation motions consist of an inferred end-effector motion. This end-effector motion is converted to a whole-body motion through the \ntwo{} mobile manipulation policy~\citeS{honerkamp2021learning, honerkamp2023learning}. This policy takes the desired end-effector motion together with the robot state and a local occupancy map for collision avoidance and produces joint velocities for the full robot at a high control frequency.

\texttt{Open object}: We provide a single demonstration for the motion of the end-effector relative to an AR marker for each articulated object.

\texttt{Pick object}: Object poses are detected via AR markers. We then define a grasp pose relative to the marker. To grasp the object, we generate a linear end-effector motion from a point in front of the grasp pose to the grasp pose and back to a point in front and slightly above the grasp pose.

\texttt{Place object}: We identify the mask of the object where the agent wants to place something down. We erode this mask and take the point closest to the robot as place pose. We predefine the height of all placeable areas to select the relevant height component. We then generate linear end-effector motions from a point in front and above the place pose to a pose above the place pose. Followed by a linear motion to the place pose and back to a point in front of the place pose.

{\footnotesize
\bibliographystyleS{IEEEtran}
\bibliographyS{bibliography}
}

\end{document}